\documentclass[utf8]{FrontiersinHarvard} 

\usepackage{url,hyperref,lineno,microtype}
\usepackage{xurl}
\usepackage[onehalfspacing]{setspace}
\usepackage{breakurl}

\def\keyFont{\fontsize{8}{11}\helveticabold }
\def\firstAuthorLast{Kamboj {et~al.}} 

\def\Address{$^{1}$School of Computing and Augmented Intelligence, Arizona State University, USA \\
$^{2}$Department of Neurology, Division of Child Neurology, University of North Carolina, USA}
\def\corrAuthor{Payal Kamboj}

\def\corrEmail{pkamboj@asu.edu}

\begin{document}
\onecolumn
\firstpage{1}

\title[The Expert’s Knowledge]{The Expert’s Knowledge combined with AI outperforms AI Alone in Seizure Onset Zone Localization using resting state fMRI.} 

\author[\firstAuthorLast ]{Payal Kamboj$^{1}$ $^{*}$, Ayan Banerjee$^{1}$, Varina L. Boerwinkle$^{2}$ and Sandeep K.S. Gupta$^{1}$ } 
\address{} 
\correspondance{} 

\extraAuth{}

\maketitle

\begin{abstract}
We evaluated whether integration of expert guidance on seizure onset zone (SOZ) identification from resting state functional MRI (rs-fMRI) connectomics combined with deep learning (DL) techniques enhances the SOZ delineation in patients with refractory epilepsy (RE), compared to utilizing DL alone. Rs-fMRI were collected from 52 children with RE who had subsequently undergone ic-EEG and then, if indicated, surgery for seizure control (n = 25). The resting state functional connectomics data were previously independently classified by two expert epileptologists, as indicative of measurement noise, typical resting state network connectivity, or SOZ. An expert knowledge integrated deep network was trained on functional connectomics data to identify SOZ. Expert knowledge integrated with DL showed a SOZ localization accuracy of 84.8 ± 4.5\% and F1 score, harmonic mean of positive predictive value and sensitivity, of 91.7 ± 2.6\%. Conversely, a DL only model yielded an accuracy of less than 50\% (F1 score 63\%). Activations that initiate in gray matter, extend through white matter and end in vascular regions are seen as the most discriminative expert identified SOZ characteristics. Integration of expert knowledge of functional connectomics can not only enhance the performance of DL in localizing SOZ in RE, but also lead toward potentially useful explanations of prevalent co-activation patterns in SOZ. RE with surgical outcomes and pre-operative rs-fMRI studies can yield expert knowledge most salient for SOZ identification.

\tiny
 \keyFont{ \section{Keywords:} Seizure onset zone, preoperative evaluation, deep learning, rs-fMRI, expert knowledge, Focal Pharmaco Resistant Epilepsy, pre surgical} 
\end{abstract}

\section{Introduction}

The World Health Organization estimates that approximately 50 million individuals worldwide are affected by epilepsy (~\cite{Hunyadi2015}). Within this population, medically refractory epilepsy (RE), constitutes about 30\%, where patients have not achieved seizure control for at least 12 months despite adequate trials of two tolerated and appropriately chosen anti-epileptic medications. RE significantly impacts the quality of life of those affected. The most successful approach for addressing RE involves surgical ablation, resection, or disconnection of brain regions associated with seizure genesis, seizure onset zone (SOZ)~\cite{Kwan2004}~\cite{Kwan2010}~\cite{Varatharajah2018}. Recent research emphasizes the importance of early diagnosis and surgical intervention to mitigate developmental complications and reduce the risk of sudden deaths ~\cite{Sillanpaa2010}. Despite advances in surgical interventions, for focal onset RE such as mesial temporal lobe epilepsy (TLE), a large number of patients (30\% – 40\%) still suffer from continued debilitating seizures  ~\cite{Boerwinkle2017} and post-surgery developmental impairments  ~\cite{Perucca2014} ~\cite{Proix2018}. Achieving a seizure-free surgical outcome is contingent upon accurately localizing the SOZ ~\cite{West2015}.\\
The gold standard technique for localization of SOZ uses invasive intracranial electro-encephalography (ic-EEG), which requires implantation of depth electrodes ~\cite{Nagahama2018}. However, concordance between ic-EEG spike density and the SOZ is observed in only 56\% of patients  ~\cite{Jobst2020}, due to sub-optimal lead implantation. As such, ic-EEG lead placement must be guided by determining the expected SOZ using non-invasive brain imaging modalities ~\cite{Chakraborty2020} ~\cite{Khoo2022}. One non-invasive method, resting state functional magnetic resonance imaging (rs-fMRI), uses blood oxygen level-dependent (BOLD) correlations measured at rest to map functional connectomics ~\cite{Smith2013}. It is an effective measure of the plasticity of large-scale networks induced by repeated and synchronized co-activation of brain regions ~\cite{GuerraCarrillo2014} caused by debilitating seizures. Independent component analysis (ICA) of rs-fMRI has been shown to have 90\% agreement with ic-EEG determined SOZs ~\cite{Boerwinkle2017} ~\cite{Jobst2020}, and the usage of rs-fMRI guided ic-EEG to locate and surgically alter the SOZ has shown significant improvement in seizure free surgical outcomes for RE pediatric patients without any increase in developmental risks ~\cite{Boerwinkle2019} ~\cite{Boerwinkle2017}.\\

One of the challenges in relating ICA’s resulting independent components (IC) of rs-fMRI to SOZ is that abnormal BOLD correlations are variable across individuals, with activation foci varying across temporal, parietal, frontal lobes, hippocampus, and cortex ~\cite{Banerjee2023}. Furthermore, over 50\% of the data comprises noise ICs, with the remaining 40-45\% attributed to resting state network (RSN), leaving only a small percentage, around 5-10\%, associated with SOZ ICs. Currently, a set of rules identified by a consortium of experts is meticulously applied with expert analysis of hundreds of ICs, rendering the pre-surgical screening process extremely time-consuming and not easily replicable ~\cite{Boerwinkle2019}. This calls for a need of a more streamlined, automated, and replicable pre-surgical screening process for SOZ localization. The existing methodology, relying on rules outlined by experts and involving detailed manual analysis of numerous ICs, is time-intensive, subjective, and lacks reproducibility. Given the demonstrated capability of large scale supervised statistical approaches, such as deep learning (DL), to identify abnormal patterns within complex datasets, recent advances have shown their application for SOZ localization from rs-fMRI data ~\cite{Nozais2021} ~\cite{Nandakumar2023}. However, a limited study on 14 subjects with refractory TLE has shown poor positive predictive value (PPV) of 52\% (± 3.9\%) for a brain parcellation to be associated with SOZ ~\cite{Nandakumar2023}. Moreover, the identified SOZs do not conform to the disease characteristics, as bilateral SOZ was identified for patients with unilateral focal TLE ~\cite{Nandakumar2023}. Our prior research has demonstrated that automating the expert rules outlined in ~\cite{Boerwinkle2019} and then implementing them in a carefully structured sequential manner result in a PPV of 65\% (±7.8\%), surpassing the performance of statistical approaches ~\cite{Banerjee2023}. \\
One of the primary reasons for poor PPV of statistical approaches is that abnormal BOLD correlations, meaning non-noise, not normal resting state fMRI network, and thus is interpreted to be pathological, form less than 10\% of the rs-fMRI ICs from ICA  ~\cite{Banerjee2023} ~\cite{Nandakumar2023}. This triggers the fundamental Achillies heel of classification science, that is class imbalance ~\cite{Branco2016}, where the statistical approach lacks enough pathological data to effectively distinguish abnormal patterns amidst significant individual variation. A pure statistical approach like DL overlooks the valuable expert knowledge encoded in terms of rules applied to real data by experts and validated by the successful seizure free outcomes upon surgical resection/ablation of expert identified SOZ. In this study, we aimed to investigate whether integration of expert knowledge of SOZ characteristics combined with data-driven statistical supervised learning approaches could improve the identification accuracy of SOZ compared to purely statistical machine learning approach, for subjects with RE. We validated identification accuracy by comparing against manual evaluation by two independent experts and subsequent ic-EEG based SOZ identification, and in surgical patients, with surgical outcomes in terms of Engel scores. Additionally, we tested which knowledge component contributed most in improving DL performance of SOZ identification. Finally, we utilized the expert knowledge contributions to generate clinically relevant explanations of SOZ identification result.


\subsection{Contribution}
This study aims to automate localization of the SOZ using DL and expert knowledge, with the primary goal of facilitating the non-invasive assessment of iEEG lead placement by the surgical team. The main contributions of the paper include:

\begin{itemize} 
\item Demonstrating that the integration of AI with expert knowledge on SOZ characteristics results in superior automated SOZ localization compared to relying solely on AI techniques for the same purpose.
\item Illustrating through knowledge ablation study that the expert knowledge of activations originating in gray matter, extending through white matter, and concluding in vascular regions, is identified as the most discriminative expert knowledge among expert-identified SOZ features.
\item Validation of SOZ localization accuracy on a large dataset of 52 patients across various age ranges and gender.

\end{itemize}

\subsection{Related works}
Recent research can be broadly categorized into two main areas, as outlined in Table 1: epilepsy detection ~\cite{Bharath2019, Lopes2012, Nguyen2021}, which entails classifying patients as either epileptic or non-epileptic based on EN identification, and SOZ localization [~\cite{Hunyadi2015, Banerjee2023, Boerwinkle2017}, the primary focus of this paper. Table 1 presents a comparative analysis of recent studies, considering factors such as the number of subjects, the proportion of the RE subgroup, age range, and the types of ICs identified. Within this domain, evaluations encompass diverse metrics such as concordance with iEEG, agreement with expert-identified SOZ, and consistency with physician assessments.

 The reported results column in Table 1 presents the evaluation metrics from the original manuscripts for each study. Various manual techniques for identifying the SOZ involve expert-defined rules based on specific spatio-temporal characteristics of BOLD signals captured by rs-fMRI. Boerwinkle et al.~\cite{Boerwinkle2017} explored the agreement between the epileptogenic zone (EZ) identified through rs-fMRI and the SOZ identified using iEEG data. They employed prevalence-adjusted bias-adjusted kappa (PABAK) on a cohort of 40 patients, revealing a concordance rate of 89\%. This study highlighted the limitations of previous approaches that focused on the most abnormal brain region for SOZ localization. However, no work is reported on automation of expert sorted ICA-based SOZ classification in this paper. Gil ~\cite{Gil2020} manually studied 21 patients with extratemporal focal epilepsy to identify SOZ related ICs in fMRI data using the general linear model-derived EEG-fMRI time courses associated with epileptic activity. Lee et al.~\cite{Lee2014} also manually investigated the functional connectivity changes in the ENs from rs-fMRI data using intrinsic connectivity contrast (ICC) to evaluate the non-invasive pre-surgical diagnostic potential for SOZ localization. The agreement of fMRI-IC with intracranial EEG SOZ was 72.4\%.

The first automation attempts were from Hunyadi et al.~\cite{Hunyadi2015}, who present a set of SOZ spatial and temporal features used to train a Least-Squares Support Vector Machine (LS-SVM). Evaluation on 18 RE patients showed sub-optimal results. DL was first explored by Nozais et al.~\cite{Nozais2021} to classify RSN ICs on non-RE patients and reported an accuracy of 92\%. However, they did not pursue SOZ identification. Luckett et al. ~\cite{Luckett2022} used 2132 healthy control data for training of 3D CNN and tested it on temporal lobe epilepsy to detect the whole hemisphere of seizure onset. The training data was synthetically altered in randomly lateralized regions which helped in detection of biological SOZ's hemisphere. Note that ICs were not used here, so this work detected the whole brain hemisphere of seizure onset rather than the brain region pointing towards the SOZ. Their primary findings suggested the ICA guided by their technique has the potential to identify epilepsy-related ICs in patients with focal epilepsy. Naresh et al. ~\cite{Nandakumar2023} explored deep graph neural networks using the T1 weighted images from rs-fMRI along with Diffusion MRI (dMRI) measurements. Study on 14 subjects showed a sensitivity of 40\% and precision of 52\% while an accuracy of 88\%. Not only the precision was sub optimal, the identified SOZs did not align with the expected disease characteristics, as bilateral SOZs were identified for patients diagnosed with unilateral focal temporal lobe epilepsy, rendering it irrelevant for pre-surgical screening.
Zhang et al.~\cite{Zhang2015} proposed ICA based automated method using unsupervised algorithm to localize the SOZ. SOZ ICs were screened based on peripheral noise IC removal, asymmetry and temporal features (excluding IC outside of frequency band 0.01-0.1hz). Consistency with the resection surgery on 10 patients was reported. If we assume consistency as true positive (TP), failure as FN and success in rejecting non-SOZ IC as true negative (TN) and failure to reject non-SOZ ICs as false positive (FP) then the results indicate significant FPs. 
Banerjee et al. ~\cite{Banerjee2023} is the most recent study in the automation of SOZ localization. It uses six expert features combined from Boerwinkle et al.~\cite{Boerwinkle2017} and  Hunyadi et al.~\cite{Hunyadi2015}. This technique reports high FPs.

\begin{table*}
\caption{Review of fMRI based IC sorting ( Not Applicable (NA), Accuracy (Acc), Sensitivity
(Sens), Specificity (Spec), Precision (Prec), Not Specified
(NS), Epilepsy networks (EN), Epileptogenic zone (EZ), M in the study column indicates manual, A indicates automation.}
\begin{center}
\scriptsize
\begin{tabular}{|p{0.7 in}|p{1.6 in}|p{0.3 in}|p{0.2 in}|p{0.5 in}|p{0.3 in}|p{1.9 in}|}
\hline
\textbf{Problem} & \textbf{Study } & \textbf{N}  & \textbf{RE} & \textbf{Age (years)}&\textbf{IC Class}  & \textbf{Reported results}  \\ 
          \hline
 Epilepsy  &Nyugen~\cite{Nguyen2021} A  & 322         
         &63         &Child (4 - 25)  & NA  & Sens= 85\%, Acc = 71\%, 
Spec = 71\%  \\ 
    \cline{2-7}
     Detection &Lopes~\cite{Lopes2012} A & 15         & 0  & Adult ($>$ 18)      & NA         & Acc = 87.5\% \\ 
     \cline{2-7}
  &Bharath~\cite{Bharath2019} A & 132        & 0 & Adult ($>$ 18)        & EN         & Sens = 100\%, Acc = 97.5\%,  Spec = 94.4\% 
 \\ 
    \hline
 SOZ &Boerwinkle ~\cite{Boerwinkle2017} M & 40       &40  &Child (1.5 - 19.8) & EZ-SOZ               & Agreement with iEEG derived SOZ = 90\%,  Prec = 79\%, Sens = 93\%   \\
   \cline{2-7}
   Localization &Gil~\cite{Gil2020} M &21         &0 &Adult ($>$ 18) &SOZ                 &NS 
 \\
  \cline{2-7}
  &Lee~\cite{Lee2014} M &29        &29 &Adult ($>$ 18) &SOZ                  &Concordance with iEEG derived SOZ = 72 \%  \\
  \cline{2-7}
   &Hunyadi~\cite{Hunyadi2015} A & 18         &18   & Adult ($>$ 18)       & SOZ         & Sens=40\%, Acc=51\%, Spec=77\% \\ 
     \cline{2-7}
     &Nozais~\cite{Nozais2021} A & 2093                 &0 & Adult ($>$ 18) & RSN   &Acc= 92\%     \\ 
   \cline{2-7}
   &Luckett ~\cite{Luckett2022} A &2164        &0 &Adult ($>$ 18) &SOZ                & Lateralization of epilepsy foci Acc = 90 \% as compared to video EEG   \\
   \cline{2-7}
  &Nandakumar~\cite{Nandakumar2023} A       & 14         & 14  & Child (9 - 18) & EZ                & Prec = 52\%, Sens = 40\%, Acc = 88\%  
 \\
    \cline{2-7}
    &Zhang~\cite{Zhang2015} A  & 10        &10 & Adult ($>$ 18)  & SOZ  & Consistency with physicians assessment   \\ 
    \cline{2-7}
    &Banerjee~\cite{Banerjee2023} A       &52         &52  &Child (0.25 - 18) &RSN, SOZ                & Prec = 93\%, Sens = 79\%, Acc = 75\%  
 \\
    \cline{2-7}    
   &\textbf{This study} A &\textbf{52}         &\textbf{52}         &Child (0.25 - 18) &\textbf{RSN, SOZ}        & Prec = 93\%, Sens = 89\%, Acc = 84\%    \\
   \hline
  \end{tabular}
  \label{LR}
\end{center}
\end{table*}

Knowledge integration into DL models has been recently explored in many domains~\cite{Cui2022,Hossain2023} including medical imaging~\cite{Xie2021} for diagnosis, lesion or organ segmentation with great success rate. Expert knowledge can be integrated in two broad ways~\cite{Cui2022}: a) scientific knowledge through mathematical models as performed in molecular dynamics analysis, or b) experiential knowledge, through logic rules. The current work falls in the second category. To the best of our knowledge, this is the first work exploring experiential knowledge integration with DL in epilepsy surgical planning.

\section{MATERIALS AND METHODS}
\subsection{Participants}
52 consecutive patients with quality data of average age 8 years 8 months (±5 years 4 months) were retrospectively studied who were diagnosed to have RE based on International League Against Epilepsy (ILAE) criteria  ~\cite{Berg2010}, from our previously published IRB approved retrospective cohort data who had ic-EEG and surgery. The evaluation involved rs-fMRI, continuous video monitoring while electroencephalography (EEG) is being performed, and anatomical 3T MRI as a part of standard MRI SOZ localization protocol followed at Phoenix Children’s hospital (PCH) for epilepsy surgery evaluation. From this published study, the three imaging modalities were independently reviewed by two blinded experts, a neurologist, and a neurosurgeon, to determine the SOZ location in both anatomical MRI and rs-fMRI. For rs-fMRI, each expert sorted independent components (ICs) into three categories: NOISE, resting state network (RSN) and SOZ. (Henceforth, class labels are denoted by capitalized, bold and italicized text). In cases where there was any disagreement, a third reviewer was consulted for the final determination. \\
Subsequently, each patient was subjected to ic-EEG based monitoring, which was independent of the rs-fMRI monitoring result. A clear indication of SOZ through observation of ic-EEG spikes determined the confirmed candidacy of the patient for surgical resection, ablation, or disconnection. For all patients, the SOZ ICs identified from rs-fMRI were manually verified by the experts. Hence, the expert manual denotation of an IC as NOISE, RSN or SOZ, supported by ic-EEG and/or surgical outcome, is considered as ground truth in this research. \\
For patients that did undergo surgery, the surgical location was determined by the expert epilepsy surgery conference team informed by the noninvasive imaging, including the expert identified rs-fMRI based SOZ location, and ic-EEG monitoring result. Validation of the manual SOZ determination was performed through evaluation of seizure free outcomes after surgical alteration of the rs-fMRI identified SOZ corroborated with ic-EEG. \\
The patients were divided into three age groups to evaluate the effect of age on the SOZ localization performance. Patients in all age groups varied across many demographic and clinical characteristics (Table 2). Surgical outcomes were evaluated using Engel scores where Engel I meant seizure free, Engel II meant at-most one debilitating seizure in the first year after surgery.  

\subsection{Data acquisition and processing}
The MRI images were obtained using a 3T MRI unit, Ingenuity Philips Medical systems, equipped with a 32-channel head-coil. The rs-fMRI settings were configured with a TR 2000 ms, TE 30 ms, matrix size 80 X 80, flip angle 80° and a total number of 46 slices. Each slice had a thickness of 3.4 mm without any gaps, and the in-plane resolution was set to 3 × 3 mm. The acquisition process involved interleaved acquisition, with a grand total of 600 volumes obtained across two 10-min runs, culminating in a total acquisition time of 20 mins. MELODIC tool ~\cite{Beckmann2004} was employed to analyze the rs-fMRI and extract ICs using ICA ~\cite{Boerwinkle2019}. Pre-processing consisted of discarding the initial 5 volumes to remove T1 saturation effects, applying a high-pass filter at 100 seconds, correcting for slice time, implementing spatial smoothing with a full-width at half maximum of 1mm, and addressing motion artifacts through MCFLIRT ~\cite{Jenkinson2002}, while excluding non-brain structures. Linear registration was done between the individual functional scans and the patient’s high-resolution anatomical scan ~\cite{Jenkinson2001} which was further refined using boundary-based registration ~\cite{Greve2009}.

\begin{table}
    \centering
    \begin{tabular}{|c|c|c|c|c|c|}
        \hline
          \textbf{Variable} & \textbf{All n = 52} & \textbf{Age 0 – $<$5,} & \textbf{Age 5 – $<$13,} & \textbf{Age 13 – 18,} \\
          & &\textbf{n = 20} &\textbf{n = 18} &\textbf{n = 14} \\
        \hline
         Sex (\% female) & 55.8\% & 65\% & 55.5\% & 43\% \\
          \hline
         Age at onset in months (s.d.) & 60 (50) & 13 (13) & 68 (36) & 118 (32) \\
          \hline
         Surgery (\%) & 48\% & 55\% & 28\% & 78.6\% \\
         Seizure Frequency at resting & & & & \\
          \hline
         state evaluation per month (s.d.) & 145 (493) & 303 (785) & 54 (92) & 37(72) \\
         Seizure frequency after & & & & \\
          \hline
         surgery (s.d.) & 4 (14) & 3 (10) & 0.5 (1) & 7 (20) \\
          \hline
         Seizure free (\%) & 64\% & 73\% & 60\% & 50\% \\
          \hline
         Ablation & 28.8\% & 25\% & 16.6\% & 50\% \\
          \hline
         Resection & 13.4\% & 15\% & 11.1\% & 14.2\% \\
          \hline
         Disconnection & 3.8\% & 10\% & 0\% & 0\% \\
          \hline
         Ethnicity- Asian & 5.7\% & 0\% & 11.1\% & 7.1\% \\
          \hline
         Ethnicity- Black/AA & 5.7\% & 5\% & 5.5\% & 7.1\% \\
          \hline
         Ethnicity- Hisp/Lat & 23\% & 10\% & 33.3\% & 28.5\% \\
          \hline
         Ethnicity- NA/Indian & 3.8\% & 10\% & 0\% & 0\% \\
          \hline
        Ethnicity- White & 59.6\% & 70\% & 50\% & 57.1\% \\
        \hline
    \end{tabular}
    \caption{Demographic and clinical characteristics of participants, including sex distribution, age at onset, surgical procedures, seizure frequencies, seizure outcomes, and ethnicity breakdown. }
    \label{tab:my_label}
\end{table}

\begin{figure*}[h!]
  \centering
  \includegraphics[width=\textwidth,height=0.8\textheight,keepaspectratio]
  {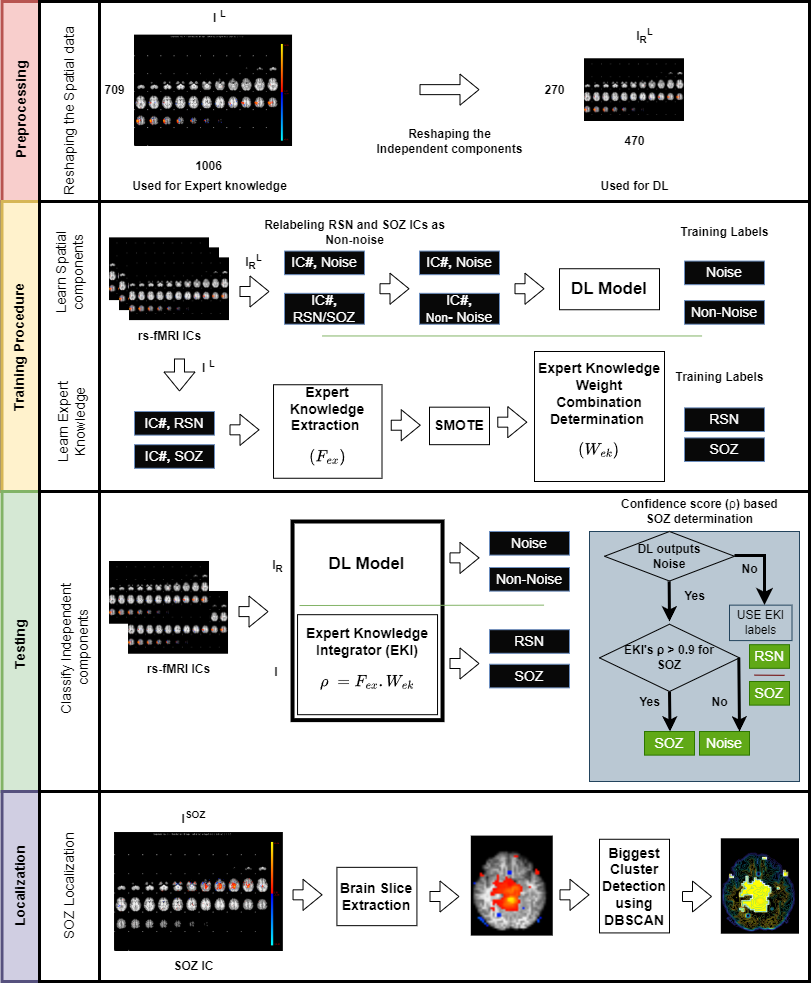}
  \caption{Overview of the proposed SOZ IC localization. Top panel: preprocessing the data by reducing the image dimensions to alleviate computational overhead. Second panel - top: training involves relabeling RSN and SOZ as non-Noise components. Second panel – bottom: These components are then subjected to CNN. Additionally, we establish an expert knowledge integration model (EKI), which is trained based on the extracted expert knowledge from RSN and SOZ components. Third panel: testing involves classification task of rs-fMRI ICs into three categories: NOISE, RSN and SOZ using both DL and expert knowledge. Bottom panel: localization of SOZ involves identification of biggest cluster amongst a patient’s SOZ slices. The operator $\circ$ denotes dot product.}
  \label{fig:example}
\end{figure*}
\subsection{Computational approach overview}
The SOZ localization approach utilized two types of models: a) deep supervised classification model or DL model, and b) expert knowledge integration (EKI) model. The approach combined the result of these two models following three steps (Figure 1): 

\textbf{Step 1 Preprocessing}: DL model used a labeled set of ICs, where each IC, $I^L$ is labeled as RSN, SOZ, or NOISE. DL model complexity drastically increases with input size potentially resulting in more data requirement to avoid under-fitting. Moreover, DL model expects all ICs to be of the same size. Hence, in the pre-processing step we resized each labeled IC, $I^L$ of size 709×1006×3, to an image $I_R^L$  of size 270×470×3. $I_R^L$ for use by the DL model, while $I^L$ was used by the EKI. $I_R^L$  gave the most optimized DL model with the best accuracy for the given dataset, determined through hyper-parameter tuning ~\cite{Goodfellow2016}. \\

\textbf{Step 2 Training}: In this phase, each RSN or SOZ IC was relabeled as $\overline{\text{NOISE}}$. The DL was trained to recognize NOISE or $\overline{\text{NOISE}}$ classes. In parallel, each RSN and SOZ IC was passed through the expert knowledge feature extraction mechanism and a weight optimization was applied to obtain the best linear combination of expert knowledge components that were most discriminative between RSN and SOZ. \\

\textbf{Step 3 Testing}: A test patient's IC, I, was passed through DL and EKI models. EKI provided a confidence score $\rho$ for I being SOZ. If DL categorized $I_R$ as $\overline{\text{NOISE}}$, I retained EKI labels. However, if DL labeled $I_R$ as NOISE, I’s classification then depended on $\rho$. Only if $\rho > 0.9$, I was marked as SOZ, else it retained DL label of NOISE. Having marked SOZ ICs, the SOZ was localized by designating the largest activation cluster, extracted using Density-Based Spatial Clustering of Applications with Noise (DBSCAN) ~\cite{Celebi2005}.    

\subsubsection{Training Phase: Noise detection using labels through supervised learning}

The ICs $I^L$ were relabeled to form $I^{L\left\{N\right\}}$ where ICs were either NOISE ICs or $\overline{\text{NOISE}}$ (RSN/SOZ). Five strategies with an 80-20 train/test split of the entire data were tested to classify $I^{L\left\{N\right\}}$ into the new class categories: a) 2D convolution neural network (CNN), b) Multilayer perceptron similar to ~\cite{Nozais2021}, c) transfer learning using VGG-16 Imagenet model  ~\cite{Talo2019}, d) problem reduction by treating the BOLD timeseries as images  ~\cite{Sobahi2022}, and e) Vision Transformer (ViT). Validation result showed that the 2D CNN had the best precision and sensitivity in determining NOISE ICs (Comprehensive results table in supplementary document). Consequently, we opted for the utilization of the 2D CNN for the classification of noise ICs. The hyperparameters of the 2D CNN were obtained using the Keras-tuner’s hyperband algorithm, with the objective of minimizing the validation loss. The hyperparameter tuning process involved exploring various configurations:
\begin{itemize}
\item Number of convolution layers: [3, 4, 5],
\item Number of units or filters per convolution layer: 32-512, with a default of 128,
\item Number of neurons in the dense layer: 192 to 1024, with a step of 256,
\item Learning rates: 0.01, 0.001, or 0.0001,
\item Dropout rates: 0.2, 0.33, 0.4, 0.5, or 0.66,

\end{itemize}
The optimized hyperparameter values, determined through the Keras-tuner, were as follows: three convolution layers, with 64, 64, and 256 3 × 3 filters in each respective layer; a dense fully connected layer with 704 neurons; a learning rate of 0.0001, and a dropout rate of 0.33. These specific hyperparameter values were selected to enhance the performance of the 2D CNN in accurately classifying noise ICs.
\begin{figure*}[h]
  \centering
  \includegraphics[width=168mm, height=120mm]{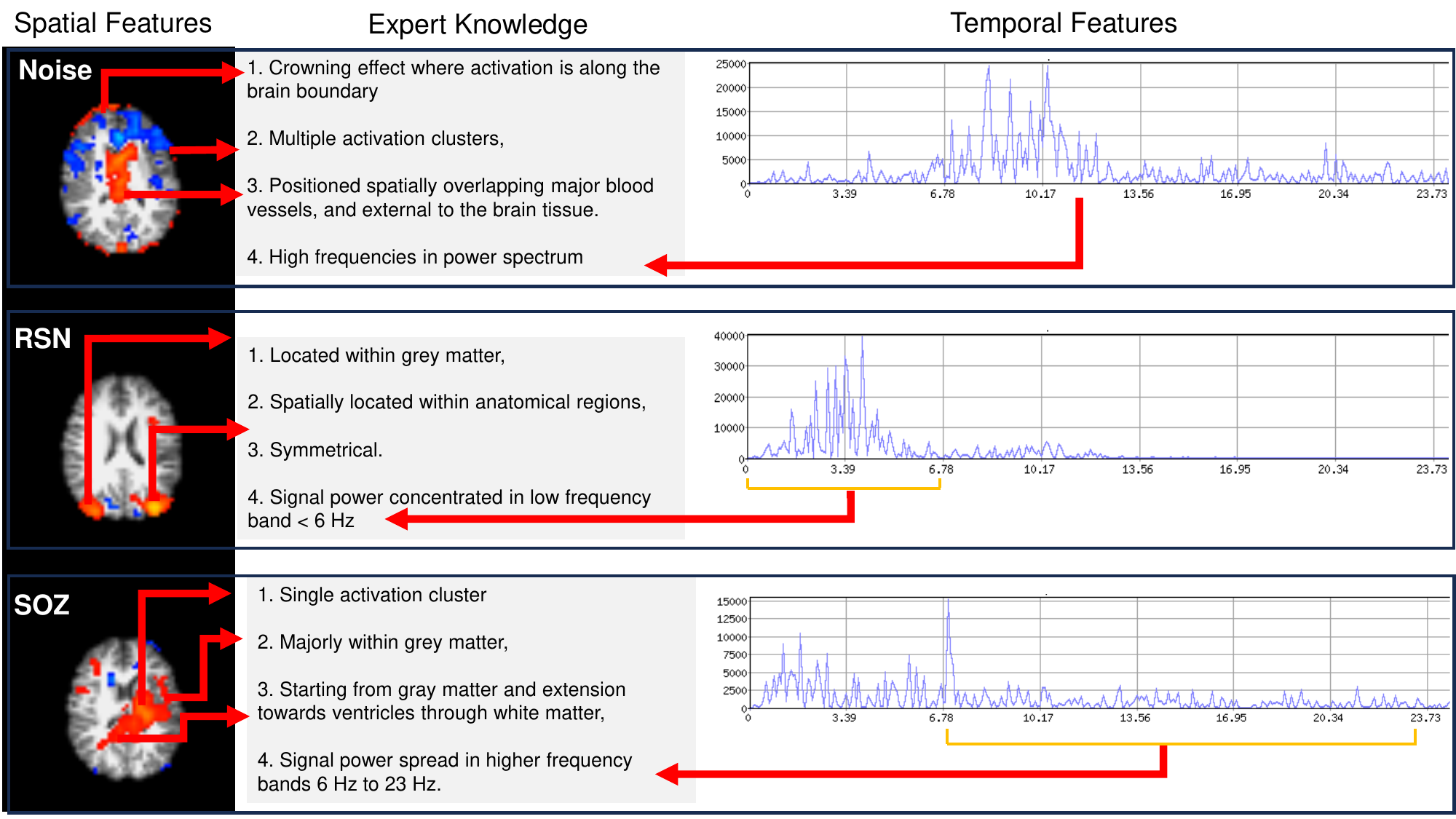}
  \caption{Three types of information are encoded in rs-fMRI: NOISE, RSN and SOZ. Each of these categories adheres to specific rules that define their classification.}
  \label{fig:example}
\end{figure*}
 Keras’s image data generator was used to create batches of both NOISE and $\overline{\text{NOISE}}$ IC images. IC images were resized from I to $I_R$ using ‘flow from directory’ method. 'Binary cross-entropy' loss function along with 'Adam' optimizer were chosen. Potential overfitting was addressed using dropout regularization and “\textit{early\_stopping}” strategy. Activation function “ReLU” was chosen for input and hidden layers, and “Sigmoid” function for the output layer. Given the characteristics of our dataset, which features dark backgrounds and required extraction of sharp features while controlling variance and computational complexity, we inserted a max pooling layer of 2 × 2 after every convolution layer  ~\cite{Goodfellow2016}. 

\subsubsection{Training Phase: Expert Knowledge on rs-fMRI IC}
Expert epileptologists use the RSN, NOISE, and SOZ indicators to manually sort the ICs (Figure 2) as compiled from the works of ~\cite{Hunyadi2015} and ~\cite{Boerwinkle2019}. In our methodology, we encoded the SOZ specific expert knowledge into the SOZ localization mechanism. This phase was subsequently divided into two steps.
\begin{figure*}[h]
  \centering
  \includegraphics[width=\textwidth, height=118mm]{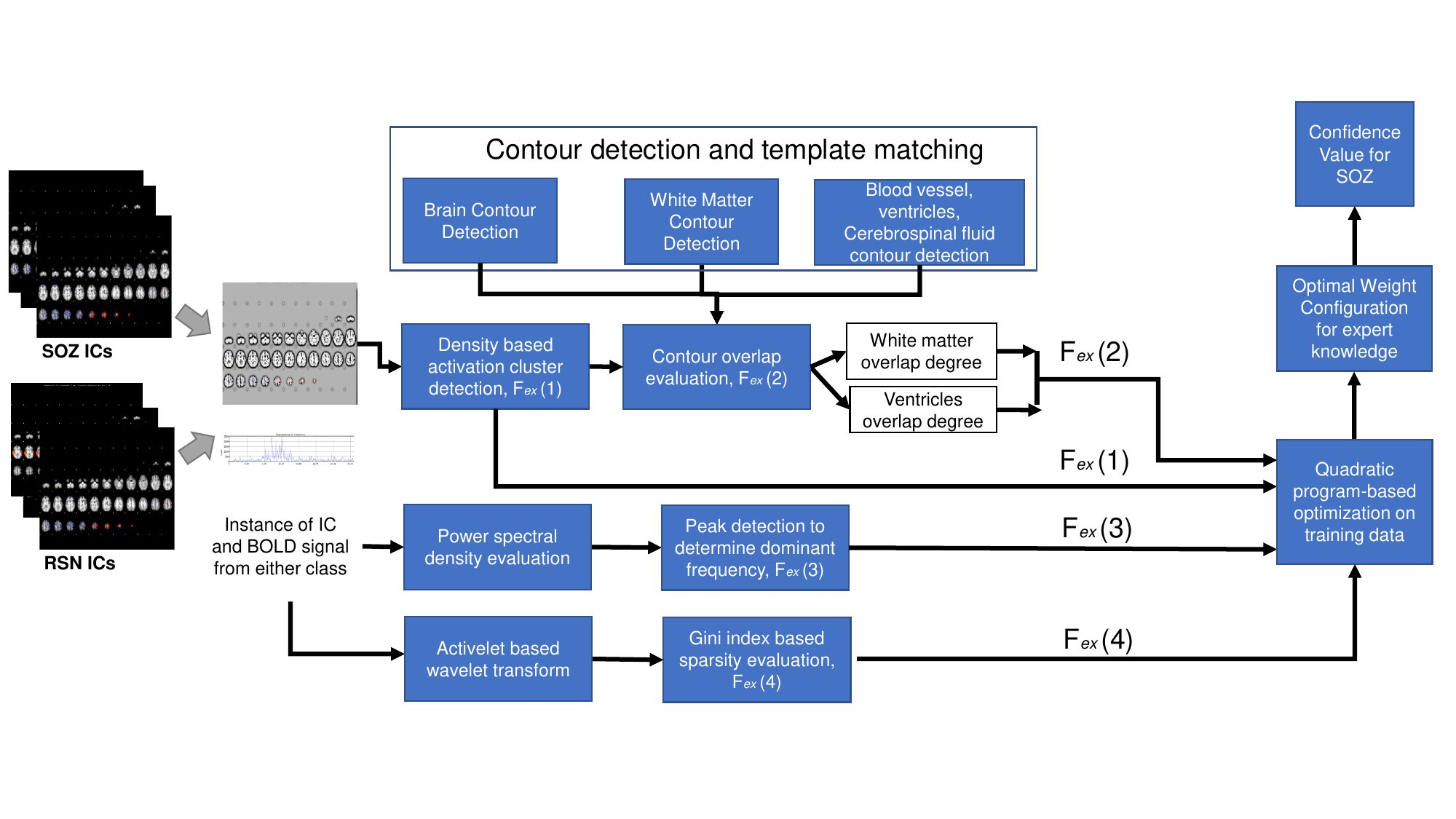}
  \vspace{-2cm}
  \caption{Expert Feature extraction and integration process.}
  \label{figEKI}
\end{figure*}

a) Extracting brain slices: Brain slices were derived from RSN and SOZ ICs through template matching. We used the Montreal Neurological Institute's 152 brain template (MNI152) for this purpose. With the help of the coordinates given by template matching, we extracted brain slices which enabled the subsequent extraction of features guided by expert knowledge.\\
b) Extraction of expert knowledge: The expert knowledge about SOZ characteristics (Figure 2) is represented using the following features, $F_{ex}$:\\

	\begin{enumerate}
    \item \textbf{$F_{ex} (1)$ Number of clusters}: SOZ IC ideally has one cluster of activation that spreads asymmetrically in one hemisphere, whereas an RSN IC consists of multiple (at least 2) clusters of activation which are spread symmetrically across the two brain hemispheres.
    \item \textbf{$F_{ex} (2)$ Activation extended to ventricles}: A SOZ has activation extended from grey matter towards ventricles through the white matter. 
    \item \textbf{$F_{ex} (3)$ Dominant frequencies}: SOZ’s BOLD signal power spectra exhibit dominant frequencies greater than 6 Hz.
    \item \textbf{$F_{ex} (4)$ Sparsity in frequency domain}: The rs-fMRI SOZ power spectrum is sparse with dominant frequency much more spread out throughout the spectrum than RSN. 
\end{enumerate}

The abovementioned features were extracted using the following method to form the feature vector $F_{ex}$ for each IC, $I^L$.

\noindent\textbf{$F_{ex}$ extraction method:}\\
\textbf{$F_{ex} (1)$ Number of clusters:} From each IC, brain slices were extracted (Figure \ref{figEKI}). From each slice, the number of clusters was estimated using DBSCAN  ~\cite{Celebi2005}. This approach had two adjustable parameters: neighborhood, which defined the distance metric and a value called $\epsilon$, and $v_{min}$, which determined the minimum number of neighboring voxels. Voxels with more than  $v_{min}$ neighbors within the  $\epsilon$ distance were considered core points and formed a cluster. Voxels that were not core points but were within  $\epsilon$ distance of a core point were classified as border points and assigned to the nearest core point's cluster. All other points were disregarded. Clusters were formed by combining core points that were within  $\epsilon$ distance of each other. Additionally, we set a threshold of 135 pixels, counting only those clusters that surpassed this threshold to determine the total number of clusters. The output of this step was the number of clusters in each IC slice (Figure \ref{figEKI}). 

\textbf{$F_{ex} (2)$ Activation extended to ventricles:} To identify the activation of the SOZ that extended from grey matter towards the ventricles through the white matter, a Sobel filter-based edge detection technique was applied, which extracted the contours for each slice, with the white matter exhibiting the most prominent contour within the slice ~\cite{Banerjee2023}. To obtain the ventricular regions, we applied edge detection to determine brain boundary. The ventricles run throughout the brain, however, they are less prominent in the slices that are towards the brain surface. Hence, we selected slices that are near the base of the brain. In these slices, the ventricle is more prominent and interrupts the continuity of the image. As such the contour detection gives multiple brain boundary contours, identified through the Sobel filter. The ventricular regions were within the convex hull of the brain boundary contours but did not intersect any brain boundary (Figure \ref{figEKI}). Subsequently, a comprehensive analysis was conducted to determine if the larger clusters (with a size exceeding 135 pixels) had the overlapping with the white matter and extension towards the ventricles. In the overlapping process, from each slice of an IC, both clusters and contours were obtained. The presence of an overlapping cluster could potentially impede the contour detection algorithm, hindering the extraction of white matter and blood vessel contours. In the initial pass through the ICs, we obtained a version of each slice devoid of clusters, serving as a basis for contour identification. The algorithm then underwent a subsequent pass through each slice of an IC, detecting clusters and evaluating their intersection with the white matter.

\textbf{$F_{ex} (3)$ Dominant frequencies} and \textbf{$F_{ex} (4)$ Sparsity in frequency domain:} For temporal SOZ characteristics, ICs were analyzed for activelet and sine dictionary sparsity in their time courses. For calculating the sparsity in activelet basis, the BOLD signal was divided into windows of length 256 samples. From every window, four levels of activelet transformation coefficients using the 'a trous' algorithm with exponential-spline wavelets were extracted  ~\cite{Hunyadi2015}. The Gini Index metric was used for activelet coefficients and sine dictionary sparsity evaluation in the frequency band of 0.01Hz to 0.1Hz.

\subsubsection{Training Phase: Balanced Dataset Creation}
The data distribution is composed of approximately 51\% NOISE, 43\% RSN and merely 5\% SOZ occurrences. To overcome class imbalance after the feature extraction process, synthetic SOZ features were created using SMOTE ~\cite{Chawla2002}. Given the constraint of a restricted quantity of available SOZ ICs, approximately 5 ICs per subject, SMOTE identified authentic SOZ IC samples within the feature space and performed linear interpolation of features.

\subsubsection{Training Phase: Expert Knowledge Combination logic}
Ambiguity is inherent in expert knowledge, given the significant individual variance in seizure onset characteristics. Hence, $F_{ex}$ could not be used in isolation to represent expert knowledge and a carefully crafted combination was necessary. We utilized the subset of ICs, $I^{L\left\{R,S\right\}}$
,that is labeled RSN or SOZ  to configure a linear combination logic for the expert knowledge vector $F_{ex}^I$ for an IC I that gave the best discriminative power between these two classes. For each $I^{L\left\{R,S\right\}}$, we defined $y_{i} = -1$ if it was RSN and $y_i=1$ if it was SOZ. We derived an expert knowledge weight vector $\omega_{ex}$, of size $|F_{\text{ex}}|$ ×1 that:

\begin{align*}
\text{Minimizes:} & \quad \sum_{i=1}^{|I^{L\left\{R,S\right\}}|} \left(1 - y_i \frac{\omega_{\text{ex}} \circ (F_{\text{ex}}^i)}{\|F_{\text{ex}}^i\|}\right)^2, \\
\text{such that:} & \quad \sum_{i=1}^{|F_{\text{ex}}|} \omega_{\text{ex}}^i = 1,
\end{align*}
where $F_{ex}^i$ is the expert knowledge vector of the $i^{th}$ IC in $I^{L{R,S}}$ , $\lVert F_{\text{ex}}^i \rVert$ is the L2 norm of a vector, and $\circ$ denotes the dot product operator (Figure \ref{figEKI}).

\subsubsection{Testing Phase: SOZ localization approach}
To obtain the robust estimate of our approach’s performance on patient’s data, we employed a leave-one-out cross-validation strategy, wherein each patient was given an opportunity to represent the entirety of the test datasets. This method of cross validation resulted in the most variance and a tight confidence interval through this approach, indicated robust performance across all patients. All rs-fMRI ICs of the test subject, I, went through a dual assessment using pre-trained DL model with $I_R$ and EKI model with I. The DL model classified $I_R$ as either NOISE or $\overline{\text{NOISE}}$. In parallel, EKI model assigned SOZ or RSN labels to the ICs I based on their confidence score $\rho_i = \frac{\omega_{\text{ex}} \circ 
(F_{\text{ex}}^I)}{\left\| F_{\text{ex}}^I \right\|}$, where $\omega_{ex}$ is the weight configuration from the training phase of the EKI model. The test IC I is assigned the label SOZ under two conditions: 

	a) $I_{R}$ was classified as $\overline{\text{NOISE}}$ by DL model but I was classified as SOZ by EKI model, or\\
 
	b) $I_{R}$ was classified as NOISE by DL model but I was classified as SOZ with a classification score $\rho> 0.9$. \\
 
Otherwise, I was not marked as SOZ. At this point, the knowledge component with the highest contribution, $\omega_{\text{ex}}^j F_{\text{ex}}^I (j)$, in determining the SOZ was subsequently highlighted as the rationale/explanation behind selecting a specific IC as the SOZ.\\

The output of this step was a set $I^{\text{SOZ}}$ of ICs $I_i^{\text{SOZ}} \in I^{\text{SOZ}}$ that were marked as SOZ. The SOZ ICs were then further processed through the brain slice extraction and DBSCAN mechanism to obtain a set of clusters $C_i \, \forall \, I_i^{\text{SOZ}} \in I^{\text{SOZ}}$. The localized SOZ was the largest cluster in each IC in $I^{\text{SOZ}}$, \\

\begin{equation}
\text{SOZ area for } I_{i}^{\text{SOZ}} = \arg \max_{j} \left\{ \left| C_{i}^{j} \right| \right\}
\end{equation}

\section{RESULTS}
We evaluated: a) efficacy of our approach, its variation across age and sex and compare with state-of-the-art techniques, b) significance of localized SOZ through correlation with surgical outcomes and c) knowledge ablation to show relative importance of spatial and temporal expert knowledge in SOZ identification.

\subsection{Comparative Techniques }
We chose the following categories for comparison with our proposed approach supervised learning with both labels and expert knowledge (SLLEK):

1) \textbf{Supervised learning with labels using CNN (SLL-CNN):} We utilized a 2D CNN-based deep learning technique for comparison, solely using the labeled dataset in a supervised manner without incorporating any form of expert knowledge encoding. We also implemented cost-sensitive learning in CNN to ensure equal significance across all three classes during gradient updates.

2) \textbf{Supervised learning with labels using ViT (SLL-ViT):} We employed another DL approach of Vision Transformer (ViT) for our comparative analysis. This methodology also relied on the labeled dataset, embracing a supervised learning paradigm without integrating any explicit encoding of expert knowledge. To optimize the model's performance on our dataset, we leveraged Optuna to identify the most effective hyperparameters. Furthermore, to address class imbalance issue within the dataset, we set the weight parameter of the loss function to the computed class weights. This is particularly crucial when certain classes are underrepresented (SOZ in our case), as it helps the model to give more emphasis to the minority classes, preventing them from being overshadowed by the majority classes. Additionally, in order to prevent ViT from suffering from gradient explosion and gradient vanishing issues, we implemented gradient clipping and batch normalization respectively. 

3) \textbf{Statistical pattern learning with expert knowledge (SLEK):} This methodology was inspired by Hunyadi et al.~\cite{Hunyadi2015} which uses expert-guided features to facilitate model learning. To ensure an unbiased comparison with our own approach, we also applied the Synthetic Minority Over-sampling Technique (SMOTE) to generate ICs endowed with SOZ features, thereby achieving balance among the three classes (implementation details in the supplementary document).

4) \textbf{Unsupervised learning with expert knowledge (ULEK):} This approach was inspired by EPIK ~\cite{Banerjee2023}, which employs a cascade of six expert rules in a waterfall technique for IC classification (detailed implementation in the supplementary documents).

\begin{table}
    \centering
    \footnotesize
    
    \begin{tabular}{|c|c|c|c|c|c|c|c|c|c|}
        \hline
          \textbf{Metrics} & \textbf{Method} & \textbf{Age} & \textbf{Age} & \textbf{Age} & \textbf{Male} &\textbf{Female} &\textbf{Overall } &\textbf{Our }\\
        &  & \textbf{0 – $<$5,} & \textbf{5 – $<$13,} & \textbf{13 – 18,} &\textbf{N = 23},  &\textbf{N = 29} &\textbf{Results}  &\textbf{Approach }\\
          & &\textbf{N = 20,} &\textbf{N = 18,} &\textbf{N = 14,} &\textbf{EOK} &\textbf{EOK} &\textbf{EOK} &\textbf{Compares}\\
          & &\textbf{EOK} &\textbf{EOK} &\textbf{EOK} & & & &\textbf{ p value}\\

        \hline
         \textbf{Accuracy}& SLL-CNN & 50\% & 38.8\% & 42.8\% &39.1\% &41.9\% &46.1\% & 0\\
          \hline
          & SLL-ViT &40\% &27.7\% &35.7\% &39.1\% &31\% &34.6\% & 0.0007 \\
          \hline
         & ULEK & 90\% & 72.2\% & 64.2\% &78.2\% &72.4\% &75\% & 0.08\\
          \hline
           &SLEK & 31.5\% &61.5\%   & 71.4\% &60.8\% &44.8\%&50\% & 0\\
        
          \hline
          &\textbf{SLLEK}  & \textbf{80\%}  & \textbf{88.8\%} & \textbf{85.7\%}  &\textbf{91.3\%} &\textbf{79.3\%} &\textbf{84.6\%}  &\textbf{NA}\\
          & &\textbf{(+30\%)} &\textbf{(+50\%)} &\textbf{(+42\%)} &\textbf{(+52\%)} &\textbf{(+37\%)}&\textbf{(+38)}&\\
          \hline
          \textbf{Precision}& SLL-CNN & 90.9\% & 100\% & 75 \%  &90\%&86.6\% &88.8\% & 0.02\\
          \hline
            & SLL-ViT &88.8\% & 100\% & 71.4\% &90\% &81.8\% &85.7\% & 0\\
          \hline
         &ULEK & 94.7\% & 100\% & 75\% & 94.7\% &91.3\%&92.8\%& 0.2 \\
          \hline
         &SLEK  & 85.7\% & 100\% & 83.3\% &93.3\%&86.6\%&89.6\%& 0.03\\
          \hline
          &\textbf{SLLEK}  &\textbf{94.1}\% & \textbf{100\%} &\textbf{85.7\%} &\textbf{95.4\%} &\textbf{95.8 \%}&\textbf{93.6\%} &\textbf{NA}\\
          &  & \textbf{(+3\%)} & \textbf{(+0\%)} &\textbf{(+10\%)} &\textbf{(+5\%)} &\textbf{(+9\%)}&\textbf{(+5\%)} &\\
          \hline
         \textbf{Sensitivity} & SLL-CNN & 52.6\% & 38.8\% & 50\% &40.9\%&44.8\%&48.9\%& 0\\
          \hline
            & SLL-ViT &42.1\% & 27.7\% & 41.6\% &40.9\% &33.3\% &36.7\% & 0\\
          \hline
         &  ULEK & 94.7\% & 72.2\% & 75\% &81.8\%&77.7\%&79.5\%& 0.1\\
          \hline
          &SLEK & 33.3\% & 61.1\% & 83.3\% &63.6\%&48.1\%&53.6\%& 0\\
          \hline
         & \textbf{SLLEK} & \textbf{84.2\%} & \textbf{88.8\%} & \textbf{100\%} &\textbf{95.4\%}&\textbf{82.1\%} &\textbf{89.7\%}&\textbf{NA}\\
         &  & \textbf{(+31\%)} & \textbf{(+50\%)} & \textbf{(+50\%)} &\textbf{(+54\%)} &\textbf{(+37\%)}&\textbf{(+41\%)} &\\
          \hline
         \textbf{F1 score} & SLL-CNN & 66.6\% & 55.9\% &60\% &55.3\%&59\%&63\%&0\\
          \hline
            & SLL-ViT & 57.1\% & 43.3\% &52.5\% &56.2\% &47.3\% &51.3\% & 0\\
          \hline
      & ULEK & 94.7\% & 83.8\% & 75\% &87.8\%&83.9\%&85.6\%&0.041\\
        \hline
       & SLEK & 47.9\% & 75.8\% & 83.3\% &76.1\%&61.8\%&67\%& 0.01\\
         \hline
       & \textbf{SLLEK} & \textbf{88.8\%} & \textbf{94\%} & \textbf{92.2\%} &\textbf{95.4\%}&\textbf{88.4\%}&\textbf{91.6\%}&\textbf{NA}\\
        &  & \textbf{(+22.2\%)} & \textbf{(+38.1\%)} &\textbf{(+32.2\%)} &\textbf{(+40.1\%)} &\textbf{(+29.4\%)}&\textbf{(+28.6\%)} &\\
        \hline
    \end{tabular}
    \caption{SOZ Identification Performance Metrics. Expert Knowledge (EoK) denotes the effect of merging expert knowledge and labels in our approach. Supervised learning with labels (SLL), Supervised learning with expert knowledge (SLEK), Unsupervised learning with expert knowledge (ULEK), Supervised learning with both labels and expert knowledge (SLLEK). }
    \label{tab:my_label}
\end{table}

\subsection{Evaluation Metrics}
We employed a two-fold approach:
a) We assessed the agreement between SOZ labelled ICs using our technique and the surgically targeted SOZ location for each Engel score group for 25 patients with available surgical resection/ablation outcomes in our dataset.
b) For all 52 patients, we validated the accuracy of generated labels from various approaches against the expert's sorted labels. The evaluation was conducted using commonly employed metrics such as accuracy, precision, and sensitivity~\cite{Hunyadi2015, Zhang2015, Lee2014, kamboj2023merging}.

\subsection{Statistical Methods}

Statistical methods were utilized to derive the significance of:
a) The effect of age and sex on the SOZ identification performance.
b) The difference in standard metrics among algorithms.

For the first aim, we utilized a mixed effects model, incorporating age and sex as predictors, along with their combined effect, and a random effect on the patient.

For the second aim, we computed the variance of the evaluation metrics across various subsets of test data obtained through categorization by age and sex. The variance of each metric in the techniques that are closest in performance to SLLEK showed $< 10\%$ difference. Moreover, we utilized the Kolmogorov-Smirnov (KS) test~\cite{karson1968handbook} to verify that the distribution of evaluation metrics across the subsets of test data came from a normal distribution with significance value $\alpha < 0.05$. The p value of the KS test is provided in Table III in supplementary document with a high p value indicating that the KS test could not reject the null hypothesis that the data came from a normal distribution. Since the variance of the evaluation metrics for the closest methods are similar and the metrics across test data subsets fit normal distribution, we utilized a one-sided t-test to evaluate the statistical significance of the difference between our approach and other comparative techniques. The 95\% confidence \( p \)-values are provided in Table 3.  

\subsection{Performance Evaluation}
Our approach (SLLEK) outperformed the other techniques across all evaluation metrics, as illustrated in Table 3 for the given patient population, warranting further investigation. The results encompassed standard metrics evaluations, considering variations in age and sex.

We observed the impact of incorporating expert knowledge with DL in our approach, quantified as the difference between SLLEK and SLL (EoK). The last column in Table 3 provides the statistical significance of the difference between SLLEK and comparative techniques implemented on our dataset.

SLLEK exhibited high sensitivity, indicating a low False Negative (FN) rate compared to other methods. Proficiency in accurately identifying the correct SOZ ICs suggests that expert knowledge integration with DL enhances the SOZ ICs identification and warrants further exploration.

SLLEK demonstrated higher accuracy, precision, and sensitivity across all age groups and sex distributions. In contrast, SLL and SLEK exhibited significant variability based on age and sex. ULEK emerged as the second-best performer after SLLEK. The \( p \)-values presented in Table 3 highlight the statistically significant differences between SLLEK and all other comparative techniques. Nevertheless, statistically, there is an insignificant difference between SLLEK and ULEK.

Comparison with state-of-the-art computer vision technique ViT is also presented in Table 3. As the results show, ViT didn't perform good for SOZ localization. This observation aligns with the understanding that ViTs may face challenges in generalizing well with smaller datasets. It's noteworthy that CNNs, in contrast, exhibit better generalization on smaller datasets, yielding better accuracy. This is attributed to the inherent capability of CNNs to excel in learning from limited data ~\cite{Mauricio2023} ~\cite{Zhu2023}.

Overall, the outcomes suggest that our approach has the potential to enhance the manual sorting workflow for the surgical team, positioning it as a promising and effective tool in detecting SOZ for pediatric RE patients.

\subsection{Performance with Surgical Outcomes} Of the 25 subjects who had surgery to remove rs-fMRI determined SOZ, 16 (64\%) achieved seizure freedom (Engel I), and 7 (28\%) experienced significantly reduced postoperative seizure frequency (Engel II). This indicated that the removed regions likely represented a substantial portion of the epileptogenic network.

\begin{table}
    \centering
    \begin{tabular}{|c|c|c|c|c|c|}
        \hline
          \textbf{Approach} & \textbf{ Ablation } & \textbf{Resection} & \textbf{Engel I} & \textbf{Engel II}  \\
        &\textbf{Procedures}   &  \textbf{Procedures} & \textbf{Outcomes} & \textbf{Outcomes}   \\
          &\textbf{(N=15)} &\textbf{(N=15)} &\textbf{(N = 16)} &\textbf{(N = 5)}  \\

        \hline
         & \textbf{Sensitivity} &\textbf{Sensitivity} & \textbf{Sensitivity} & \textbf{Sensitivity}  \\
          \hline
         SLL& 66.6\% & 57.1\% & 56.2\% & 80\%   \\
          \hline
           ULEk &33.3\% & 42.5\% &43.7\%   & 60\%  \\
        \hline
           SLEK &73.3\% & 71.4\% &75\%   & 100\%  \\
           \hline
           \textbf{SLLEK}  &\textbf{93.3\%} &\textbf{85.7\%} &\textbf{93.7\%}   & \textbf{100\%}  \\

          \hline
       
        \hline
    \end{tabular}
    \caption{Performance comparison of Methods across surgical procedures and Engel outcomes. Supervised learning with labels (SLL, Supervised learning with expert knowledge (SLEK), Unsupervised learning with expert knowledge (ULEK), Supervised learning with both labels and expert knowledge (SLLEK).}
    \label{tab:my_label}
\end{table}

SLLEK showed the highest sensitivity of 93.3\% for patients undergoing minimally invasive ablation surgery, making it a promising option in such cases. For patients undergoing resection, SLLEK maintained a consistent sensitivity of 85.7\%, outperforming other techniques for the dataset.

Furthermore, when analyzing patients with Engel I outcome, SLLEK exhibited a 93\% agreement with expert sorting, reinforcing its suitability and reliability as a pre-surgical screening tool (Table 5).
\subsection{Knowledge Ablation Studies}

SLLEK’s performance could be attributed to the influence of each expert knowledge component on the accuracy of SOZ identification. To better understand its capabilities, we assessed the impact of removing specific knowledge components (Table 5) from SLLEK in relation to standard metrics used in Table 3. 

\begin{table}
    \centering
    \begin{tabular}{|c|c|c|c|c|}
        \hline
          \textbf{Metrics} & \textbf{ Accuracy } & \textbf{F1 score } & \textbf{Change in F1 score}   \\

        \hline
       
         SLLEK without activelet sparsity& 84.6\% & 91.6 \% & 0\%   \\
          \hline
           SLLEK without sine sparsity &84.6\% & 91.6 \% &0\%    \\
        \hline
           SLLEK without nos. clusters &75\% & 85.8\% &$\downarrow$ 6.3\%   \\
           \hline
          SLLEK no white matter overlap   &\textbf{50\%} &\textbf{66.6\%} &$\downarrow$\textbf{27.4\%}  \\

          \hline
       
        \hline
    \end{tabular}
    \caption{SLLEK Knowledge ablation study (Accuracy, F1 score, and change in overall F1 score). (Supervised learning with both labels and expert knowledge (SLLEK)).}
    \label{tab:my_label}
\end{table}

\textbf{SLLEK without temporal features:} The BOLD signal temporal features were removed one by one from the expert knowledge model of SLLEK. We created two unique configurations: a) SLLEK without activelet domain sparsity, and b) SLLEK without sine domain sparsity. Table 5 reveals no significant impact on metrics, indicating that removing temporal features had limited effect on the classification of patient's ICs with SLLEK.

\textbf{SLLEK without spatial features:} The spatial features were removed one by one from the expert knowledge model to create two unique configurations: a) SLLEK without the number of clusters, and b) SLLEK without white matter overlap. An 11\% reduction in accuracy and a 6\% drop in F1 score were noted when the number of clusters feature was removed from the analysis. However, when the white matter overlap feature was omitted, a substantial 41\% decrease in accuracy and a 27\% reduction in the F1 score were observed. These findings underscore the pivotal role of white matter overlap as the most influential feature in the identification of the SOZ.

\section{DISCUSSIONS and LIMITATIONS}

 The results suggest an approach that combines expert features and AI for SOZ localization may possess the capability to generate connectivity classifications that align with ic-EEG and surgical outcomes. This stems from the design where expert knowledge integration model facilitates the derivation of weight contributions for each expert feature for SOZ identification. This provision not only enables explanations for the selection of an IC for SOZ but also amplifies its potential as an advanced tool for SOZ identification in clinical contexts. By furnishing transparent rationales for its classifications, our approach may equip the surgical team with invaluable insights.

The interplay between the deep supervised classification model and expert knowledge integration components in the SOZ localization approach is instrumental in achieving superior localization accuracy. The DL model excels in discerning noise images, as evidenced by its classification capabilities. Our PCH dataset of 52 patients had a total of 5616 IC images where only 5.6\% of these images represented SOZ ICs, while 51.1\% were attributed to Noise ICs, and 43.1\% to RSN ICs. Due to such high data imbalance, where majority class is 16.6 times more prevalent than minority class, commonly used imbalanced data handling techniques such as cost sensitive training failed to provide good performance as seen in SLL-CNN, or SLL-ViT. Similarly under-sampling the majority classes to balance the data would have resulted in significant information loss from the majority class, potentially resulting in overall performance loss.

Due to balanced data between Noise and RSN ICs, DL could learn their distribution. However, due to the limited availability of SOZ ICs in the dataset, traditional DL techniques faced challenges in learning the intricate features of these rare events from such a small subset of SOZ data. To overcome this limitation, a need arose for a methodology that could leverage the wealth of expert knowledge on SOZ characteristics available in the literature review. Additionally, relying solely on expert knowledge also exhibits a sub-optimal outcome as it struggles to capture the intricate details of brain networks, possibly because of the overlapping characteristics between Noise and SOZ ICs. For instance, an activation located in the white matter is associated with Noise, whereas an activation originating in the grey matter, extending into the white matter and reaching the ventricles, is indicative of SOZ. The minimal overlap of activation on grey matter can sometimes make these SOZ activations appear as a noise. This is a domain where DL excels in encoding the nuances of Noise ICs more effectively, benefiting from a slightly larger data of Noise ICs to learn and represent their characteristics. This unique integration of DL for noise IC classification and EKI for SOZ IC classification addresses the performance limitations inherent in relying solely on either DL or EKI strategies, offering a more robust and comprehensive solution to SOZ IC identification.

While our dataset is one of the largest in recent literature for pediatric patients with RE, a larger study is necessary to address the potential impact of variability in fMRI preprocessing and motion correction techniques, which can differ across centers. Before being used with minimal expert supervision, further testing of this technique in real-world settings is necessary, considering its intended application in local epilepsy care centers. 

\section{CONCLUSION}
The most effective treatment for RE is surgical resection or ablation of the SOZ which requires accurate localization to avoid functional brain network damage and developmental impairments. While rs-fMRI, a non-invasive imaging technique, holds promise for SOZ localization and guiding iEEG lead placement, its clinical integration is hindered by the lack of expertise in manual seizure onset analysis. Additionally, manual sorting of ICs obtained from rs-fMRI data using ICA is a challenging and subjective task, as only a small fraction (less than 5\%) of the ICs is related to the SOZ. This makes the process time-consuming and limits the reproducibility and availability of this non-invasive technique. Accurate, automated and reproducible SOZ localization is imperative for successful surgical treatment of RE while avoiding functional brain network damage and resultant developmental impairments. This study shows how expert knowledge can be integrated with powerful supervised learning approaches to automate SOZ localization. Reliable performance on a large dataset of children with RE, stratified across age, sex, and corroboration with one-year post-operative Engel outcomes for rs-fMRI guided surgery increases confidence of potential for clinical integrability of the approach. The Activations initiating in gray matter, extending through white matter and ending in vascular regions were seen as the most discriminative expert identified SOZ characteristics.  The prospect of automating SOZ localization using advanced AI techniques and existing expert knowledge not only addresses existing challenges in manual analysis but also suggests a transformative shift towards more accessible, trustworthy and reproducible clinical application in epilepsy care. In the future, a multi-center study to evaluate general applicability of the technique irrespective of scanning protocols and measurement devices is contemplated.

\section*{Conflict of Interest Statement}
None of the authors have any conflict of interest to disclose.

\section*{Author Contributions}

Payal Kamboj was responsible for designing, implementing, and statistical analysis of the proposed methodology, as well as for writing the paper.  Ayan Banerjee played a significant role in implementing the expert knowledge feature extraction and made equal contribution to the paper’s writing. Varina Boerwinkle was responsible for interpretation of rs-fMRI data, explaining Noise, RSN and NOISE biomarkers, and thoroughly reviewing the paper writing with respect to epilepsy care. Sandeep Gupta provided valuable insights related to the proposed methodology and validated the credibility of this research.

\section*{Acknowledgments}
We are thankful to Phoenix Children Hospital, Sarah Wyckoff and Bethany Sussman for rs-fMRI data collection, and subject de-identification. The authors did not receive a grant from any organization for this work.

\section*{Data Availability Statement}
The code is available at \url{https://gitfront.io/r/user-9511950/nAmPjJ6L7WyQ/SOZ-Localization-using-DL-and-expert-knowledge/}. Since the dataset includes a pediatric population, access to it requires a data sharing agreement and ethics review by Phoenix Children's Hospital (PCH).

\bibliographystyle{Frontiers-Harvard} 


\end{document}